\documentclass[11pt,a4paper]{article}

\usepackage[utf8]{inputenc}
\usepackage[T1]{fontenc}
\usepackage{lmodern}
\usepackage[margin=2.5cm]{geometry}
\usepackage{amsmath}
\usepackage{amssymb}
\usepackage{pifont}
\usepackage{graphicx}
\usepackage{booktabs}
\usepackage{array}
\usepackage{tabularx}
\usepackage{xcolor}
\usepackage{authblk}
\usepackage[backend=biber,style=numeric-comp,sorting=none,giveninits=true,maxbibnames=99,texencoding=ascii]{biblatex}
\usepackage{url}
\usepackage[hidelinks]{hyperref}

\graphicspath{{./}}
\newcolumntype{L}{>{\raggedright\arraybackslash}X}
\newcolumntype{P}[1]{>{\raggedright\arraybackslash}p{#1}}

\title{From Hosted Service to Field-Owned Infrastructure: Open, Self-Hosted Real-Time Camera-Trap Alerting with Regional Open-Weight Models}
\author[ ]{Paul Fergus}
\affil[ ]{ Liverpool John Moores University, Byrom Street, Liverpool L3 3AF, United Kingdom}
\affil[ ]{Correspondence: \href{mailto:p.fergus@ljmu.ac.uk}{p.fergus@ljmu.ac.uk}}

\date{}

\begin{document}
\maketitle

\begin{abstract}
Camera traps connected to cellular networks can alert field teams within minutes of capturing an image, a capability used against poaching, to protect threatened species and to mitigate human--wildlife conflict. Most such systems have run inference on central servers, so every image, including incidental images of people and time-stamped records of where high-value species are, leaves the organisation that collected it and often the country in which it was taken. We ask whether the same capability can run on infrastructure that field organisations own. We describe Trap Tracker RT, an open-source, self-hosted application that receives images from cellular cameras through a mailbox or, without leaving the site network, by FTP/SFTP, classifies them with a project-specific model and sends alerts through a mail server of the organisation's choosing. Using DeepFaune-SSA v1.2, which pairs MegaDetector v6 with a DINOv3-based classifier for 33 sub-Saharan taxa, as a case study, we compared it with three openly available alternatives on 31,131 out-of-distribution images from Kenya, Namibia and Malawi that are independent of its training data. DeepFaune-SSA achieved an image-level species accuracy of 0.901 (95\% CI 0.894--0.908), against 0.530 for the global SpeciesNet classifier and 0.556 and 0.547 for single-stage YOLOv10x and YOLO26x detectors trained on the same regional data, and it had the highest recall for all 26 species evaluated. The single-stage detectors failed to find an animal in about a quarter of images, whereas SpeciesNet found animals but named the species in only 57\% of images, falling back to coarser labels in most of the rest. In bench tests of the fully local path, alerts were issued about 6~s after an upload completed, including over slowed and interrupted uploads, and the two-stage pipeline processed 19 images per minute on eight CPU threads without a graphics processor. Trap Tracker RT is in use by five organisations in the United Kingdom and Kenya, including for human--elephant conflict mitigation. We discuss how data paths, rather than the location of inference, determine data sovereignty, the responsibilities that self-hosting transfers to field organisations, and the ethical questions that real-time monitoring raises.

\medskip
\noindent\textbf{Keywords:} camera traps; real-time alerting; edge computing; data sovereignty; self-hosting; open-weight models; species classification; domain shift; sub-Saharan Africa; conservation technology
\end{abstract}

\section{Introduction}

Camera traps are now a standard tool for monitoring terrestrial mammals~\parencite{oconnell2011,burton2015,wearn2019}, and the volume of images they produce has grown rapidly~\parencite{kays2020}, making manual review a substantial burden~\parencite{glover2019}. Deep learning has largely removed that constraint: models can now locate and identify animals with accuracy approaching that of trained annotators in many settings~\parencite{norouzzadeh2018,willi2019,tabak2019,tuia2022}. In most applications, however, this automation is retrospective. Memory cards are collected after weeks or months and the images are processed in batches, which suits ecological inference but not management decisions that depend on timely information. Rhinoceros and pangolin have been heavily poached across sub-Saharan Africa~\parencite{biggs2013,ingram2018}, and patrol planning benefits from knowing where illegal activity is taking place~\parencite{critchlow2015,fang2016}. Camera traps that transmit images over cellular or satellite networks have therefore been used to alert rangers and wildlife authorities to people, target species and large carnivores within minutes of capture~\parencite{whytock2023,dertien2023,fergus2023welgevonden,chalmers2025}.

How such systems are built matters as much as the models they use. A common design hosts inference centrally. Conservation AI, a platform developed by the author's group and since retired, received images by email from cellular cameras, classified them on GPU servers in the United Kingdom and returned alerts to partners in several countries~\parencite{fergus2024consai,vsd2023}. Central hosting concentrates computation and maintenance where hardware and expertise are available, but every image leaves the site at which it was captured and, usually, the country. Camera traps routinely photograph people~\parencite{sandbrook2018}, whose images are personal data under laws that restrict their transfer abroad, and time-stamped images of high-value species from identified cameras disclose where those animals are~\parencite{lindenmayer2017,cooke2017}. Central hosting also makes field operations depend on the infrastructure and funding of a single institution, as the retirement of Conservation AI illustrates, and it repeats a pattern in which data generated in African countries are processed and retained elsewhere~\parencite{abebe2021,birhane2020}.

An alternative is to run inference on a computer that the field organisation owns, receiving images from ordinary cellular cameras over the site network. This position lies between hosted services and on-camera edge devices, which avoid transmitting images but require specialised hardware and compact models that recognise few classes~\parencite{zualkernan2022,whytock2023,dertien2023}. It is useful only if the models an organisation can run identify species at sites they were not trained on, which camera-trap models often fail to do~\parencite{beery2018,schneider2020,koh2021}. Three kinds of openly released model are now available for sub-Saharan Africa. Global classifiers such as SpeciesNet, trained on tens of millions of images, cover more than 2,000 labels~\parencite{gadot2024,google2025}. Regional single-stage detectors, such as the YOLOv10x and YOLO26x models released by the author's group for 34 African taxa~\parencite{fergus2026yolossa}, locate and name animals in a single pass. Two-stage pipelines pair a general-purpose animal detector such as MegaDetector~\parencite{beery2019,hernandez2024} with a regional classifier fine-tuned from an open-weight foundation model~\parencite{rigoudy2023,simeoni2025,fergus2026shoulders}. These approaches have not been compared on the same out-of-distribution African images, so a field organisation choosing a model for alerting has little evidence on which to base the choice.

This paper continues our work on open camera-trap AI. We have released single-stage detectors for UK and sub-Saharan African mammals~\parencite{fergus2026,fergus2026yolossa} and the self-hostable Trap Tracker platform. We have also shown that open-weight classifiers can be adapted to UK species with modest data and compute~\parencite{fergus2026shoulders}. Here we turn to deployment. We describe Trap Tracker RT, an open-source, self-hosted application~\parencite{traptracker2026} that accepts images from cellular cameras through a mailbox or an entirely local file-transfer path, applies a project-specific single-stage or two-stage model, and sends alerts through a mail server chosen by the organisation. As a case study we use DeepFaune-SSA v1.2~\parencite{fergus2026ssa}, a two-stage pipeline that pairs MegaDetector v6 with a DINOv3-based classifier for 33 sub-Saharan taxa, and compare it with SpeciesNet and with the regional YOLOv10x and YOLO26x detectors on 31,131 images from three countries that contributed nothing to its training data.

Two research questions guide the study. The first concerns the system: can a self-hosted application on field-owned hardware issue alerts within seconds of an upload, including over slow and interrupted cellular uploads and without a graphics processor (RQ1)? The second concerns the models: how does a regional two-stage open-weight pipeline compare with a global classifier and with regional single-stage detectors at sites that contributed nothing to its training (RQ2)? RQ1 determines whether local hosting is practical, but RQ2 determines whether the alerts it produces are worth receiving, since an animal that is missed raises no alert at all.

In addressing these questions, the paper makes three contributions. First, it presents the design of Trap Tracker RT and shows that whether images leave a site depends on the routes by which they arrive and alerts depart, not on where inference runs. Second, it reports bench measurements of alert latency under slow and interrupted uploads, and of throughput on GPU and CPU. Third, it provides an out-of-distribution benchmark of four openly available models on sub-Saharan camera-trap images from Kenya, Namibia and Malawi, with confidence intervals that resample whole camera sites.

The remainder of the paper is organised as follows. Section~\ref{sec:related} reviews automated camera-trap analysis, real-time and edge systems, and the sensitivity of camera-trap data. Section~\ref{sec:design} describes the system, Section~\ref{sec:methods} the models and evaluation protocols, and Section~\ref{sec:results} the results. Section~\ref{sec:deployments} summarises earlier hosted deployments, Section~\ref{sec:discussion} discusses implications and limitations, and Section~\ref{sec:conclusion} concludes.

\section{Related Work}
\label{sec:related}

Early work showed that convolutional neural networks could identify species in camera-trap images at near-human accuracy when trained on large labelled collections~\parencite{tabak2019}, such as Snapshot Serengeti~\parencite{swanson2015,norouzzadeh2018,willi2019}. A persistent difficulty is generalisation to new locations. Beery, Van Horn and Perona showed that classifiers generalise poorly to camera sites not seen in training, and may rely on the backgrounds of the cameras they were trained on~\parencite{beery2018}; Schneider et al.\ found that accuracy fell substantially at locations not seen in training~\parencite{schneider2020}; and Shepley et al.\ and Norman et al.\ examined how far models generalise across locations and habitats~\parencite{shepley2021,norman2023}. Camera-trap imagery is now a reference case of distribution shift in machine learning~\parencite{koh2021}. Out-of-distribution evaluation at unseen sites, rather than random hold-out splits, is therefore needed before a model is used operationally.

This sensitivity to site has shaped current architectures. MegaDetector, a class-agnostic detector for animals, people and vehicles trained on millions of images from many projects~\parencite{beery2019,hernandez2024}, has been found effective at separating images containing animals from empty images, and at flagging images of people, in independent assessments~\parencite{fennell2022,leorna2022,velez2023,mitterwallner2024}. It decouples the broadly transferable problem of finding an animal from the region-specific problem of naming it, and classifying detector crops rather than whole images improves species accuracy~\parencite{gadot2024}. SpeciesNet follows this design at global scale: it combines MegaDetector with a classifier covering more than 2,000 labels, filters predictions by the country in which an image was taken, and returns a genus, family or higher label when uncertain~\parencite{gadot2024,google2025}. Regional classifiers built on the same principle include DeepFaune for European fauna~\parencite{rigoudy2023} and DeepFaune-UK~\parencite{fergus2026shoulders}, which rest on vision transformers~\parencite{dosovitskiy2021} pretrained with self-supervision~\parencite{oquab2024,simeoni2025} that can be fine-tuned with modest labelled data. Single-stage detectors from the YOLO family~\parencite{redmon2016,wang2024yolov10,ultralytics2025yolo26} offer a simpler alternative that localises and names animals in one pass and is cheaper to run, but must learn both tasks from the regional data alone. Desktop applications that run such models, or review their output, offline~\parencite{whytock2021,mbaza,vanlunteren2023,greenberg2019} and cloud platforms that host data and models together~\parencite{ahumada2020} have made them usable by ecologists, and comparative evaluations recommend weighing workflow, data management and software requirements alongside accuracy~\parencite{velez2023}.

Reviews of conservation technology describe rapid advances in sensors, networked devices and automated data processing~\parencite{lahozmonfort2021,bergertal2018,stephenson2020,speaker2022}. Existing real-time camera-trap systems differ in where inference takes place and in the route by which images and detections travel (Table~\ref{tab:systems}).

\begin{table}[htbp]
\centering
\caption{Real-time and offline camera-trap AI systems, classified by where inference is performed and how images reach it.}
\label{tab:systems}
\small
\begin{tabularx}{\textwidth}{@{}P{3.4cm}LLL@{}}
\toprule
System & Where inference runs & How images arrive & Real-time \\
\midrule
Conservation AI (retired)~\parencite{fergus2024consai} & Hosted servers, United Kingdom & Email from 3G/4G cameras & Yes \\
Gabon smart camera~\parencite{whytock2023} & Edge (Raspberry Pi bridge) & Label sent over Iridium satellite & Yes, median 7.35~min \\
TrailGuard AI~\parencite{trailguard,dertien2023} & Edge (on-camera processor) & Cellular, radio or satellite & Yes, about 30~s (tigers and people) \\
Instant Detect 2.0~\parencite{instantdetect} & Base station and cloud & LoRa, then Iridium & Yes \\
Sentinel~\parencite{sentinel} & Edge retrofit & Satellite or cellular & Near real-time \\
Wildlife Insights~\parencite{ahumada2020} & Hosted (cloud) & Batch upload & No \\
Mbaza AI~\parencite{whytock2021,mbaza}, AddaxAI~\parencite{vanlunteren2023} & Offline desktop & Local folders & No \\
Trap Tracker RT (this work) & Computer owned by the field organisation & Email, or FTP/SFTP to the local computer & Yes, about 6~s after upload (bench) \\
\bottomrule
\end{tabularx}
\end{table}

At one extreme, inference runs on or beside the camera. Embedded processors have been used to classify images or filter out empty frames on the device~\parencite{nazir2017,elias2017,zualkernan2022,cunha2021}. In Gabon, Whytock et al.\ attached a single-board computer to a camera and transmitted only labels over the Iridium satellite network, with a median latency of 7.35 minutes in an area without cellular coverage~\parencite{whytock2023}, and TrailGuard AI detects tigers and people on a processor inside the camera, with notifications arriving about 30~s after the camera is triggered~\parencite{dertien2023}. Such systems minimise transmission, which matters where bandwidth is scarce, but bind users to particular hardware and to models small enough to run on it. At the other extreme, hosted platforms such as Conservation AI receive full images from commodity cellular cameras and classify them on central servers~\parencite{fergus2024consai,fergus2023welgevonden,chalmers2025}, which permits large models but sends every image off site; the authors of that platform identified edge processing as future work~\parencite{fergus2024consai}. Edge computing more generally places computation close to the source of data to reduce latency, bandwidth and exposure to third parties~\parencite{shi2016,satya2017}. To our knowledge, no existing system combines commodity cellular cameras, regional species-level models and inference on hardware owned by the field organisation with a path on which images need not leave the site.

The images such systems carry are sensitive in two ways. First, camera traps capture people as a matter of course: over 90\% of researchers surveyed by Sandbrook et al.\ reported that their projects had photographed people, usually unintentionally~\parencite{sandbrook2018}, and Sharma et al.\ proposed a code of conduct for handling such images~\parencite{sharma2020}. Images of identifiable people are personal data in each of the four countries in which the hosted deployments discussed in this paper took place, and each regulates their transfer abroad (Table~\ref{tab:law}).

\begin{table}[htbp]
\centering
\caption{Data-protection law governing the transfer of personal data abroad in the four deployment countries.}
\label{tab:law}
\small
\begin{tabularx}{\textwidth}{@{}P{2.2cm}P{4.3cm}L@{}}
\toprule
Country & Law & Transfer provision \\
\midrule
South Africa & Protection of Personal Information Act 2013~\parencite{popia2013} & s72: transfer abroad only with comparable protection, consent, or listed exceptions \\
Kenya & Data Protection Act 2019~\parencite{kenyadpa2019} & ss48--50: appropriate safeguards proven to the Data Commissioner; s50 allows processing to be required in Kenya \\
Uganda & Data Protection and Privacy Act 2019~\parencite{ugandadppa2019} & s19: equivalent protection or consent; s1(b) extends the Act to processors abroad \\
United Kingdom & UK GDPR~\parencite{ukgdpr} & Chapter V (Articles 44A--50, as amended by the Data (Use and Access) Act 2025) \\
\bottomrule
\end{tabularx}
\end{table}

Second, images of animals reveal where they are. Lindenmayer and Scheele documented cases in which published locations of rare species were exploited by poachers and collectors~\parencite{lindenmayer2017}, Tulloch et al.\ proposed a framework for weighing the risks of publishing biodiversity data~\parencite{tulloch2018}, and Cooke et al.\ described animal-tracking data being used to locate and kill tracked animals~\parencite{cooke2017}. A stream of time-stamped images of rhinoceros or pangolin from identified cameras is, in effect, a live map of where those animals are. These concerns sit within a wider debate in which camera traps and other sensors have been analysed as instruments of surveillance that extend to people living around protected areas~\parencite{arts2015,adams2019,duffy2019}, and in which data governance is pulled between open reuse under the FAIR principles~\parencite{wilkinson2016} and the authority of communities over data that concern them under the CARE principles and Indigenous data sovereignty~\parencite{carroll2020,kukutai2016,hummel2021}. Sandbrook et al.\ set out principles for the socially responsible use of conservation monitoring technology~\parencite{sandbrook2021}. For real-time camera-trap systems these debates have a concrete technical correlate: the route by which each image travels, and the organisations and jurisdictions through which it passes, is fixed by the system's architecture.

\section{System Design}
\label{sec:design}

\begin{figure}[htbp]
\centering
\includegraphics[width=\textwidth]{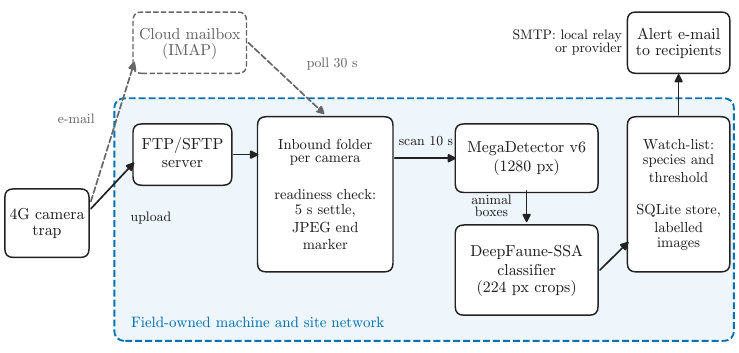}
\caption{Trap Tracker RT architecture. On the local path (solid arrows), cameras upload by FTP/SFTP to the field organisation's own computer, and alerts leave through an SMTP server of the organisation's choosing; no image needs to leave the site. The cloud-mailbox path (grey, dashed) accepts any camera that can send email, at the cost of passing every image through the mail provider. The model shown is the two-stage DeepFaune-SSA pipeline; a single-stage YOLO model can replace both model boxes.}
\label{fig:architecture}
\end{figure}

The considerations reviewed above suggest four properties that a real-time system should have if it is to be operated by field organisations under their own control. It should work with commodity cellular cameras rather than specialised hardware; it should offer a path on which no image need leave the site network; it should allow each organisation to select the model appropriate to its region and to replace it as better models become available; and it should operate on hardware that such organisations can realistically obtain, which in many settings means a laptop without a discrete graphics processor. Trap Tracker RT was designed around these properties. It is a containerised web application that a field organisation runs on its own computer. Each project has its own cameras, model, watch-list and alert recipients, and a background process for each project converts incoming images into alerts (Figure~\ref{fig:architecture}).

\subsection{Image Ingestion}

Two ingestion routes feed a common per-project inbound folder containing one subfolder per camera. In the first, cameras email each image to a mailbox, which the system polls over IMAP (by default every 30~s); messages whose subject carries a registered camera token are retained, their image attachments saved and the messages deleted. This route works with any camera able to send email, but every image passes through the servers of the mail provider. In the second, cameras upload directly to an FTP or SFTP server on the same computer, which writes into the corresponding camera subfolder. Mailbox ingestion can be disabled for a project, which then requires no email account at all and keeps every image within the site network.

Uploads over weak cellular links can take tens of seconds, and a file that is read before its transfer is complete will be misclassified. The system therefore scans the inbound folder every 10~s and processes a file only once its size has been stable for 5~s and, for JPEG images, once the end-of-image marker (\texttt{FF D9}) has been received. A JPEG that never completes is processed after 300~s so that a failed upload cannot block the queue. The need for this check was established empirically: an early version read a 40~s upload one second after it began, found no animal, and silently failed to raise an alert (Section~\ref{sec:op-results}).

\subsection{Models}

Models are discovered from a designated folder when the application starts, and each project selects one. Two families are supported. Single-stage detectors from the YOLO family, including YOLOv10 and YOLO26~\parencite{redmon2016,wang2024yolov10,ultralytics2025yolo26}, locate and name species in a single pass and are paired with their class lists. Two-stage pipelines first apply a general detector such as MegaDetector v6~\parencite{beery2019,hernandez2024} at its native resolution of 1280~px to locate animals, people and vehicles, and then apply a species classifier to a padded square crop of each animal box, following the evidence that cropped classification outperforms whole-image classification~\parencite{gadot2024}. The classifier's input size, normalisation, crop padding and class list are read from metadata embedded in the model file, and boxes labelled as people or vehicles retain the detector's label so that they can also trigger alerts.

Inference uses ONNX Runtime~\parencite{onnxruntime}. Because a mismatch between a model and its class list produces labels that are confidently wrong rather than obviously broken, the application refuses a class list whose length differs from the class names embedded in the model, and it never substitutes a different model when the selected one is unavailable.

\subsection{Watch-Lists and Alerts}

Each project specifies the species on which to alert and a minimum confidence for each (0.50 by default). When a processed image contains a watch-listed label at or above its threshold, every recipient receives an email containing the annotated image and the original, once per rule and image. Mail can be sent through any SMTP server, including a commercial provider with authentication or a relay on the local network without it. The choice of relay determines whether alerts, and the images attached to them, leave the site.

\subsection{Implementation and Licensing}

The default build runs on any CPU; an optional build uses ONNX Runtime with CUDA and falls back to the CPU if the GPU is unavailable. Original images, annotated copies and a SQLite database of detections are retained on the local computer, and projects that were running resume automatically after a restart. The web interface has no authentication and stores mailbox credentials locally, so it is bound to the local host and is not intended to be exposed to a network. The software is released under the Trap Tracker Research and Non-Commercial Licence~\parencite{traptracker2026}. MegaDetector v6 is distributed under AGPL-3.0 and the DINOv3 backbone under Meta's DINOv3 licence~\parencite{dinov3license}, so organisations redistributing a pipeline must observe both; the DeepFaune-SSA v1.2 weights are released under CC BY-NC 4.0~\parencite{fergus2026ssa}.

\section{Models and Evaluation Methods}
\label{sec:methods}

\subsection{DeepFaune-SSA v1.2}

DeepFaune-SSA v1.2 identifies 33 sub-Saharan taxa and is published with open weights~\parencite{fergus2026ssa} and with its training and evaluation code~\parencite{deepfaunessa2026repo}. It was chosen as the case study because it is representative of the regional models that a field organisation would deploy in a self-hosted system. The pipeline uses MegaDetector v6 (MDv6-yolov10-e at 1280~px)~\parencite{hernandez2024,pytorchwildlife,wang2024yolov10} without modification to locate animals, people and vehicles, followed by a classifier in the DeepFaune family~\parencite{rigoudy2023} built on a DINOv3 ViT-L/16 backbone~\parencite{simeoni2025,dosovitskiy2021} and fine-tuned at 224~px on square crops padded by 20\% of the longer side of each box. Because people and vehicles are handled by the detector, they were excluded from the classifier's training data.

The training set comprised 52,292 images of 33 taxa, drawn from sources including several Snapshot Safari projects~\parencite{pardo2021,swanson2015}, WCS Nigeria, and from African partners over the last 7 years. Images were partitioned into training, validation and test sets by site, so that images from a given camera appear in only one partition. Training used an inverse-frequency sampler, cross-entropy loss with label smoothing of 0.1~\parencite{szegedy2016} and mixed-precision arithmetic. The released checkpoint was that with the highest validation macro-F1 (0.995); no external data were used for model selection.

\subsection{Comparison Models}
\label{sec:baselines}

DeepFaune-SSA was compared with three openly available models that a field organisation in sub-Saharan Africa might otherwise choose (Table~\ref{tab:models}).

\textbf{SpeciesNet} (package 5.0.5, model 4.0.3a)~\parencite{gadot2024,google2025} was run with its default ensemble of MegaDetector v5a and the SpeciesNet crop classifier, with geographic filtering set to the country of each dataset (KEN, NAM or MWI), exactly as in the corresponding evaluation of DeepFaune-UK~\parencite{fergus2026shoulders}. Its image-level prediction was mapped to the DeepFaune-SSA vocabulary at the rank each class names: the species for species classes, the genus for genus classes such as genet and baboon, and the family for rhinoceros. Labels at coarser ranks, such as ``bovidae family'', ``mammal'' or ``animal'', which SpeciesNet returns when uncertain, were counted as incorrect, since they would not trigger an alert for a watch-listed species. SpeciesNet is distributed as a PyTorch package rather than in the ONNX form that Trap Tracker RT loads, so it was run with its own software.

\textbf{YOLOv10x and YOLO26x} (v1.0.0)~\parencite{fergus2026yolossa} are single-stage detectors for 34 sub-Saharan taxa plus people and vehicles, released by the author's group and trained on the same regional dataset as DeepFaune-SSA, so that the comparison isolates the architecture rather than the training data. Both are supported directly by Trap Tracker RT and were run through its inference code at the applications' default input size of 640~px, which matches their training resolution; YOLOv10x scored lower at 1280~px on a 3,000-image subset (0.518 against 0.556). Their boxes were scored in the same way as DeepFaune-SSA's, with the detector confidence of each box serving as its confidence.

\begin{table}[htbp]
\centering
\caption{Models compared on the out-of-distribution evaluation set.}
\label{tab:models}
\small
\begin{tabularx}{\textwidth}{@{}P{2.7cm}P{4.2cm}LP{2.0cm}@{}}
\toprule
Model & Architecture & Training data & Licence \\
\midrule
DeepFaune-SSA v1.2~\parencite{fergus2026ssa} & MegaDetector v6 + DINOv3 ViT-L classifier & 52,292 regional images, 33 taxa & CC BY-NC 4.0 (weights) \\
SpeciesNet 4.0.3a~\parencite{google2025} & MegaDetector v5a + EfficientNetV2-M classifier, geofenced & More than 65 million images, more than 2,000 labels & Apache 2.0 \\
YOLOv10x v1.0.0~\parencite{fergus2026yolossa} & Single-stage YOLOv10x, 640~px & Same regional data, 34 taxa plus person and vehicle & CC BY-NC 4.0 \\
YOLO26x v1.0.0~\parencite{fergus2026yolossa} & Single-stage YOLO26x, 640~px & As YOLOv10x & CC BY-NC 4.0 \\
\bottomrule
\end{tabularx}
\end{table}

\subsection{Out-of-Distribution Evaluation Protocol}
\label{sec:ood-methods}

Given the evidence that camera-trap classifiers degrade at unseen locations~\parencite{beery2018,schneider2020,norman2023}, the models were evaluated on three public datasets from LILA BC that contribute no images to its training data and are therefore out of distribution (Table~\ref{tab:ood-data}). Snapshot Safari and WCS data were excluded because the training set draws on them. Images with a single label mapping to a DeepFaune-SSA class were sampled, with a cap of 1,000 images per class per dataset and a fixed random seed. Three labels denote close relatives of a DeepFaune-SSA class (Cape porcupine, vervet monkey and an unspecified jackal) and were scored only at genus level. No evaluation image matched a training image by SHA-256 hash; a perceptual-hash search identified 7,042 candidate near-duplicates, none of which survived a pixel-correlation test and visual inspection.

\begin{table}[htbp]
\centering
\caption{Out-of-distribution evaluation datasets (LILA BC, CDLA-Permissive-1.0).}
\label{tab:ood-data}
\small
\begin{tabularx}{\textwidth}{@{}LP{4.0cm}rr@{}}
\toprule
Dataset & Country and habitat & Images used & SSA species \\
\midrule
Biome Health Project Maasai Mara 2018~\parencite{lilamara} & Kenya, savanna & 10,803 & 22 \\
Desert Lion Conservation~\parencite{liladesertlion} & Namibia, desert & 9,272 & 13 \\
Nkhotakota Camera Traps~\parencite{lilankhotakota,appel2025} & Malawi, miombo woodland & 11,056 & 16 \\
\bottomrule
\end{tabularx}
\end{table}

The protocol followed that used for the corresponding evaluation of DeepFaune-UK~\parencite{fergus2026shoulders}. Each image received a single label: for DeepFaune-SSA and the YOLO models, the species assigned to its most confident animal box with detector confidence of at least 0.25, or empty if there was none; for SpeciesNet, its own image-level prediction. The same images, clusters and bootstrap resamples were used for every model. Accuracy and macro-F1 were computed at species and genus level. Because images from the same camera are not independent, 95\% confidence intervals were obtained from 1,000 bootstrap resamples~\parencite{efron1979} of whole clusters~\parencite{field2007}, defined as camera sites or, for the Desert Lion dataset, which lacks site identifiers, as image sequences; there were 5,632 clusters in total.

\subsection{Operational Evaluation Protocol}

Operational behaviour was assessed on a workstation with an NVIDIA RTX A6000 GPU, an Intel Core i9-13900K CPU (24 cores, 32 threads) and 128~GB of memory, running Trap Tracker RT in an isolated container with DeepFaune-SSA v1.2. A local FTP server (pyftpdlib) wrote into the project's inbound folder and a local SMTP server received the alerts; mailbox ingestion was disabled and the relay required no authentication, so that no image or alert left the machine. A script emulated a camera by uploading images, drawn from the out-of-distribution evaluation set and correctly labelled by the model, under four conditions: a first image immediately after start-up, a steady 40~s upload emulating a weak cellular link, an upload paused for 20~s mid-transfer, and a 40~s upload processed by the version of the system that preceded the readiness check.

Throughput was measured as the time taken by the processing loop for each image, comprising file reading, detection and classification, on 120 images from the out-of-distribution evaluation set (40 per dataset), discarding a single warm-up image. Configurations included the GPU, all 32 CPU threads and eight CPU threads; the eight-thread runs were pinned to four performance cores with simultaneous multithreading as an approximation to a laptop-class processor. The two-stage DeepFaune-SSA pipeline and the single-stage YOLOv10x and YOLO26x models were timed. YOLO26x was timed later on a second sample of 120 images drawn in the same way; YOLOv10x, re-timed on that sample, gave the same results to within 0.01~s, and because single-stage cost does not depend on image content the two samples are comparable.

\section{Results}
\label{sec:results}

\subsection{In-Distribution Performance}

On the held-out test partition (6,865 crops), the classifier alone achieved a macro-F1 of 0.991. When the full pipeline was run on the test images, with MegaDetector finding the animals and the classifier naming them, and each predicted box matched to an annotated box, it achieved a precision of 0.887, a recall of 0.951 and an F1 of 0.916. Of the 904 false positives, 854 were detections without a corresponding annotated box; many were unannotated animals or partial views.

\subsection{Out-of-Distribution Performance}
\label{sec:ood}

On the out-of-distribution datasets, DeepFaune-SSA v1.2 achieved an image-level species accuracy of 0.901 (95\% CI 0.894--0.908) and a macro-F1 of 0.902 (Table~\ref{tab:ood}). Accuracy varied little between desert, savanna and woodland (0.894 to 0.913), suggesting that the model does not rely on the backgrounds of any one habitat, and it was comparable to that reported for DeepFaune-UK on its own out-of-distribution evaluation (0.888 for v1.2 and 0.906 for v1.3)~\parencite{fergus2026shoulders}. The detector failed to find the animal in 1.6\% of images.

\begin{table}[htbp]
\centering
\caption{Image-level results of DeepFaune-SSA v1.2 on the out-of-distribution datasets. ``Exact'' scores the species; ``genus level'' also credits the SSA congener for the three genus-only labels. 95\% confidence intervals from a cluster bootstrap over camera sites.}
\label{tab:ood}
\small
\begin{tabular}{@{}lrccc@{}}
\toprule
Subset & Images & Accuracy (95\% CI) & Macro-F1 (95\% CI) & Animal found \\
\midrule
All, exact species & 27,678 & 0.901 (0.894--0.908) & 0.902 (0.886--0.910) & 0.984 \\
All, genus level & 31,131 & 0.903 (0.896--0.909) & 0.903 (0.888--0.911) & 0.984 \\
Desert Lion, exact & 8,728 & 0.913 (0.905--0.920) & 0.882 (0.863--0.939) & 0.982 \\
Maasai Mara, exact & 9,894 & 0.894 (0.884--0.905) & 0.922 (0.900--0.931) & 0.980 \\
Nkhotakota, exact & 9,056 & 0.897 (0.880--0.911) & 0.922 (0.901--0.933) & 0.991 \\
\bottomrule
\end{tabular}
\end{table}

The most frequent errors were between morphologically similar taxa, often in night-time infrared images: impala classified as common duiker (93 images), confusion between vervet monkey and baboon (90 and 54), spotted hyena classified as jackal (90), aardvark as giant pangolin (76), and eland as bongo (68). Seven classes could not be evaluated because they do not occur in these datasets: gorilla, chimpanzee, giant pangolin, kob, bongo, giant pouched rat and rabbit. Representative detections and errors are shown in Appendix~\ref{app:examples}.

\subsection{Comparison with Other Models}
\label{sec:comparison}

\begin{figure}[htbp]
\centering
\includegraphics[width=\textwidth]{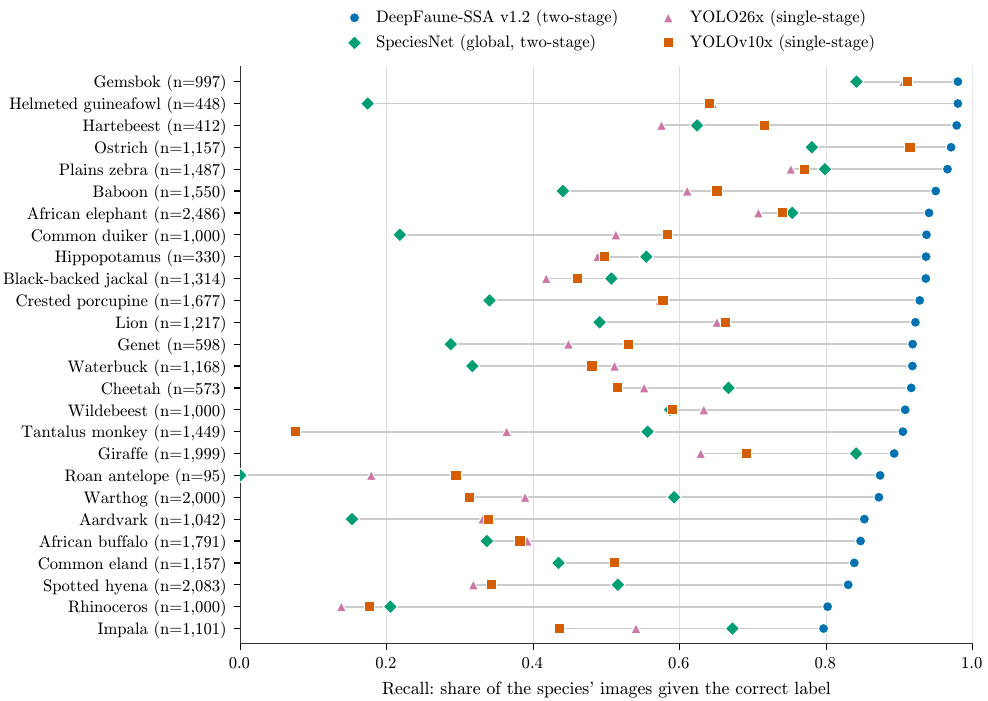}
\caption{Per-species recall of the four models on the out-of-distribution datasets (genus level, so the vervet and Cape porcupine images count towards their SSA congeners, shown as tantalus monkey and crested porcupine). Species are ordered by DeepFaune-SSA recall; $n$ is the number of test images. For SpeciesNet, labels coarser than the species (for example ``bovidae family'') count as incorrect.}
\label{fig:recall}
\end{figure}

All three alternative models were substantially less accurate than DeepFaune-SSA on the same images (Table~\ref{tab:compare}). SpeciesNet achieved an accuracy of 0.530 (95\% CI 0.503--0.557), YOLOv10x 0.556 (0.539--0.574) and YOLO26x 0.547 (0.529--0.566), against 0.901 for DeepFaune-SSA, and the confidence intervals did not overlap in any dataset. DeepFaune-SSA had the highest recall for every one of the 26 species (Figure~\ref{fig:recall}). The newer YOLO26x architecture did not improve on YOLOv10x: their accuracies were within each other's confidence intervals, and YOLO26x had higher recall for 10 of the 26 species.

\begin{table}[htbp]
\centering
\caption{Image-level species accuracy (exact species, 95\% CI) and the share of images in which an animal was found, for the four models on the same images and protocol. Macro-F1 for all images in the last row.}
\label{tab:compare}
\footnotesize
\setlength{\tabcolsep}{4pt}
\begin{tabular}{@{}lcccc@{}}
\toprule
Subset & DeepFaune-SSA v1.2 & SpeciesNet & YOLOv10x & YOLO26x \\
\midrule
All & 0.901 (0.894--0.908) & 0.530 (0.503--0.557) & 0.556 (0.539--0.574) & 0.547 (0.529--0.566) \\
Desert Lion & 0.913 (0.905--0.920) & 0.630 (0.617--0.645) & 0.643 (0.629--0.658) & 0.619 (0.604--0.635) \\
Maasai Mara & 0.894 (0.884--0.905) & 0.614 (0.586--0.640) & 0.542 (0.514--0.569) & 0.555 (0.527--0.585) \\
Nkhotakota & 0.897 (0.880--0.911) & 0.342 (0.303--0.381) & 0.488 (0.449--0.525) & 0.469 (0.433--0.502) \\
\midrule
Animal found (all) & 0.984 & 0.977 & 0.766 & 0.755 \\
Macro-F1 (all) & 0.902 & 0.606 & 0.617 & 0.613 \\
\bottomrule
\end{tabular}
\end{table}

The models failed in different ways. The two single-stage detectors failed mainly to find animals: about a quarter of images contained no box at confidence 0.25 or above (for YOLOv10x, including 847 spotted hyena, 624 warthog and 614 porcupine images; for YOLO26x, 919 spotted hyena, 652 porcupine and 624 giraffe images), compared with 1.6\% for DeepFaune-SSA. Where they did detect an animal, both named it correctly in 72\% of images. SpeciesNet, whose MegaDetector stage found an animal in 97.7\% of images, failed instead to name it. It returned a species-level label for only 57\% of images, and when it did so the label was correct in 93.6\% of cases. In a further 31.7\% of images it returned a coarser label that was correct but uninformative for alerting, such as ``bovidae family'' for a buffalo, ``mammal'' or ``animal''; in 7.8\% it returned a coarser label that was wrong, such as ``cervidae family'' for an African antelope; and the remaining 4\% were blank, unclassified or labelled as people. SpeciesNet was weakest in the Malawian woodland (0.342) and for species such as roan antelope, aardvark, helmeted guineafowl, rhinoceros and common duiker, whose recall was 0.22 or lower. Examples of each failure mode, with the four models' labels side by side, are shown in Appendix~\ref{app:examples} (Figures~\ref{fig:A1}--\ref{fig:A5}).

\subsection{Alert Latency on the Local Path}
\label{sec:op-results}

Table~\ref{tab:bench} reports the bench results. The first image after start-up was delayed by approximately 70~s while the GPU runtime selected its algorithms, a cost incurred once per start. Thereafter, alerts were delivered approximately 6~s after an upload completed, both for a steady 40~s upload and for an upload interrupted for 20~s. This delay is governed by the 5~s settling interval and the phase of the 10~s scan cycle, so it would be expected to lie between approximately 5 and 15~s; GPU inference contributed well under one second. Without the readiness check, the slow upload was read one second after it began, no animal was found and no alert was issued.

\begin{table}[htbp]
\centering
\caption{Bench results for the fully local path (FTP in, local SMTP relay out, GPU build).}
\label{tab:bench}
\small
\begin{tabularx}{\textwidth}{@{}P{4.2cm}P{2.8cm}LP{3.0cm}@{}}
\toprule
Scenario & Upload & Result & Alert after upload finished \\
\midrule
First image after start & 0.35~s & Giraffe, 0.92 & 70~s (GPU warm-up) \\
Slow upload, weak 4G & 40~s, steady & Zebra, 0.93 & 6~s \\
Stalled upload & 20~s pause halfway & Wildebeest, 0.94 & 6~s \\
Slow upload, before the readiness check & 40~s, steady & Read 1~s in: no animal found, no alert & Missed \\
\bottomrule
\end{tabularx}
\end{table}

\subsection{Throughput with and without a GPU}

\begin{figure}[htbp]
\centering
\includegraphics[width=0.85\textwidth]{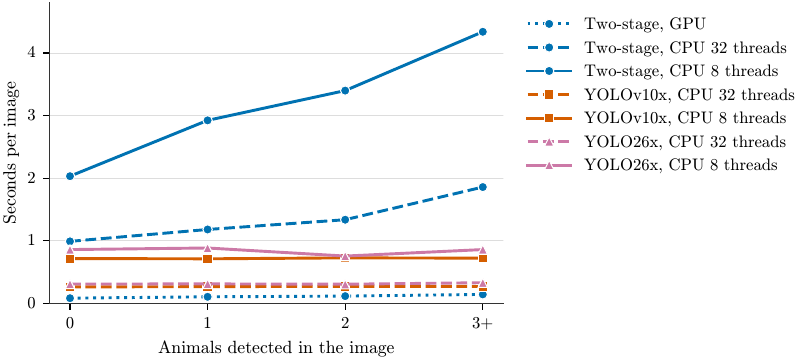}
\caption{Mean time per image against the number of animals detected in it (images with three or more animals pooled). The two-stage pipeline classifies each animal box, so its cost grows with the number of animals; the single-stage models' does not.}
\label{fig:latency}
\end{figure}

Table~\ref{tab:throughput} reports processing time per image. On eight CPU threads, the two-stage pipeline took a mean of 3.18~s per image (19 images per minute); on 32 threads it took 1.29~s, and on the GPU 0.11~s. Because the pipeline classifies each detected animal separately, its cost increased with the number of animals, from 2.0~s for images without animals to 4.3~s for images with three or more on eight threads (Figure~\ref{fig:latency}). The single-stage models' cost was independent of image content and three to four times lower on the CPU: 0.72~s per image for YOLOv10x and 0.85~s for YOLO26x on eight threads.

\begin{table}[htbp]
\centering
\caption{Time per image in the Trap Tracker RT processing loop (file read, detection and classification), 119 timed images after one warm-up image. YOLO26x was timed on a second sample of the same size (Section~\ref{sec:methods}).}
\label{tab:throughput}
\small
\begin{tabular}{@{}llccrc@{}}
\toprule
Hardware & Model & Mean & 95th percentile & Images/min & First image \\
\midrule
GPU (RTX A6000) & Two-stage DeepFaune-SSA & 0.11~s & 0.15~s & 542 & 58~s \\
CPU, 32 threads & Two-stage DeepFaune-SSA & 1.29~s & 1.83~s & 47 & 1.4~s \\
CPU, 8 threads & Two-stage DeepFaune-SSA & 3.18~s & 4.69~s & 19 & 3.2~s \\
CPU, 32 threads & Single-stage YOLOv10x & 0.27~s & 0.29~s & 225 & 0.3~s \\
CPU, 8 threads & Single-stage YOLOv10x & 0.72~s & 0.89~s & 84 & 0.8~s \\
GPU (RTX A6000) & Single-stage YOLO26x & 0.08~s & 0.12~s & 744 & 30~s \\
CPU, 32 threads & Single-stage YOLO26x & 0.31~s & 0.33~s & 194 & 0.3~s \\
CPU, 8 threads & Single-stage YOLO26x & 0.85~s & 1.17~s & 70 & 1.1~s \\
\bottomrule
\end{tabular}
\end{table}

Combining these figures with the settling and scanning intervals, the expected delay from upload completion to alert on an eight-thread CPU is approximately 8 to 20~s. The CPU build also avoids the GPU's start-up delay, which may matter at quiet sites where the first image after a restart is the one of interest. The eight-thread configuration approximates, but does not reproduce, a laptop processor; low-power laptops and single-board computers would be expected to be slower.

\section{Real-Time Deployments}
\label{sec:deployments}

The motivation for this work derives from real-time deployments in four countries that used the hosted Conservation AI platform before its retirement (Table~\ref{tab:deployments}). These deployments used YOLO-family models on central servers, and the outcomes reported below are therefore not results for Trap Tracker RT or DeepFaune-SSA. They are included because they document the value that field partners obtained from real-time alerts and the operational conditions, including weak signal and loss of equipment, under which any successor system must work.

\begin{table}[htbp]
\centering
\caption{Real-time deployments of the hosted Conservation AI platform (since retired). Camera numbers and outcomes not given in the cited sources are from the project records of the author's group.}
\label{tab:deployments}
\footnotesize
\begin{tabularx}{\textwidth}{@{}P{1.5cm}P{3.4cm}P{2.0cm}LL@{}}
\toprule
Country & Site and partners & Target & Scale and period & Outcome reported \\
\midrule
South Africa & Welgevonden Game Reserve, Limpopo; Gap Africa Projects; Vodacom~\parencite{fergus2023welgevonden} & Rhinoceros and other large mammals & 27 Reolink 4G cameras, May 2022 to Feb 2023; 37,900 images (18,520 with detections) & 12 species detected; weak 3G/4G signal; cameras destroyed by elephants \\
Uganda & Murchison Falls National Park; Chester Zoo, Uganda Wildlife Authority~\parencite{chesterzoo} & Giant pangolin & About 10 Reolink 4G cameras; project since 2018, Murchison Falls since 2022 & Email alert within 4 minutes of a pangolin detection \\
Kenya & Mount Kenya Wildlife Conservancy; Chester Zoo, Kenya Wildlife Service~\parencite{chesterzoobongo} & Mountain bongo & About 40 cameras from April 2024 & Alerts used by rangers to locate and tag newborn calves \\
UK & 11 curlew nests, Wales; GWCT Wales; Reolink; Vodafone~\parencite{chalmers2025} & Curlew and chicks & 1,072 images analysed & Accuracy 91.2\%; curlew F1 0.951, chick F1 0.960 \\
UK & Knowsley Safari~\parencite{fergus2024consai} & Paddock animals & Early deployment, still collecting in 2024 & Not reported \\
\bottomrule
\end{tabularx}
\end{table}

Two features of these deployments bear directly on the present work. First, the field conditions were demanding: cellular coverage was intermittent and cameras were damaged by elephants~\parencite{fergus2023welgevonden}, and theft and vandalism are a widespread problem for camera trapping~\parencite{meek2019} and among the barriers practitioners cite most often~\parencite{glover2019}. Stolen cameras continued to transmit images of the people who had taken them, and case reports from Uganda and the United Kingdom describe poaching recorded on camera and leading to prosecutions; these reports have not been made public for security reasons. Second, every one of these images, including those of people, was processed on servers in the United Kingdom, irrespective of where it was captured. Each deployment therefore required organisational approval under a strict protocol: images had to be held securely and could not be shared without the prior permission of the partner organisation, and images containing people were blurred or removed, depending on the organisation, as were other identifying details such as vehicle number plates. No partner refused such approval outright, but several organisations decided not to adopt the platform after initial meetings. Their reasons were not always stated, and the transfer of images out of the country may have been one of them.

Trap Tracker RT is now used by five organisations (Table~\ref{tab:rt-deployments}). In the United Kingdom, the Animal and Plant Health Agency (APHA), part of the Department for Environment, Food and Rural Affairs, has used it since 2024 to monitor grey squirrels with about ten cameras, running a single-stage YOLOv10x model and later the YOLO26x UK mammals detector~\parencite{fergus2026}; this work has not been published because of sensitivities surrounding grey squirrel management. GWCT and the two Kenyan sites replaced the hosted platform with Trap Tracker RT and use commodity Reolink 4G cameras. In Wales, GWCT, a partner of the author's group for about seven years, has used Trap Tracker RT with about 60 cameras since early 2026 to monitor farm gates and curlew nests, continuing the work described above. It used a YOLOv10x model for UK mammals over that period and now uses the YOLO26x UK mammals detector~\parencite{fergus2026}. Also in Wales, the author's group and Fieldmark AI, which develops land-management software for farms and estates~\parencite{fieldmark}, have run a biodiversity study at Maes Truan Farm since July 2026, using 13 Reolink 4G cameras and the two-stage DeepFaune-UK pipeline~\parencite{fergus2026shoulders}, the only current deployment to use a two-stage model; close to 50,000 images have been processed so far. In Kenya, the Friends of Karura Community Forest Association has used about ten cameras since February 2026 to monitor wildlife in Karura Forest, Nairobi, and Project PARC has run three cameras since September 2026 for human--elephant conflict mitigation: when an elephant is detected, the alert e-mail sent to the project's mail server is retrieved from the mailbox and used to open a valve on the water supply to a well that is shared with the local community. Both Kenyan sites run the single-stage YOLO26x sub-Saharan Africa model~\parencite{fergus2026yolossa}. The PARC deployment uses an alert to act on the environment rather than only to inform people, which makes the latency and reliability of the local path an operational requirement rather than a convenience. Both Kenyan sites currently run a single-stage model; in the light of Section~\ref{sec:comparison}, they are candidates for switching to the two-stage DeepFaune-SSA pipeline, which Trap Tracker RT supports without other changes. Beyond these real-time deployments, the companion offline application, Trap Tracker Offline, which runs the same models on folders of images that have already been collected, is used by the People's Trust for Endangered Species, the British Hedgehog Preservation Society, MammalWeb, the Zoological Society of London, Natural England and Hogwatch, a hedgehog monitoring programme. Because both applications run on the organisations' own computers, images processed are not reported centrally and total volumes across all users are unknown; the author's group's own deployments alone have processed more than one million images in the last 18 months.

\begin{table}[htbp]
\centering
\caption{Current deployments of Trap Tracker RT.}
\label{tab:rt-deployments}
\footnotesize
\begin{tabularx}{\textwidth}{@{}P{1.2cm}P{3.0cm}P{1.6cm}P{1.6cm}P{2.6cm}L@{}}
\toprule
Country & Organisation & Since & Cameras & Model & Purpose \\
\midrule
UK & Animal and Plant Health Agency (APHA), Defra & 2024 & About 10 & YOLOv10x, then YOLO26x~\parencite{fergus2026}, UK mammals & Monitoring grey squirrels \\
UK & Fieldmark AI, Maes Truan Farm, Wales & Jul 2026 & 13 Reolink 4G & DeepFaune-UK, two-stage~\parencite{fergus2026shoulders} & Farm biodiversity study; close to 50,000 images processed \\
UK & GWCT Wales & Early 2026 & About 60 Reolink 4G & YOLO26x, UK mammals~\parencite{fergus2026}; previously YOLOv10x & Monitoring farm gates and curlew nests \\
Kenya & Friends of Karura Community Forest Association, Nairobi & Feb 2026 & About 10 Reolink 4G & YOLO26x, sub-Saharan Africa & Wildlife monitoring \\
Kenya & Project PARC & Sep 2026 & 3 Reolink 4G & YOLO26x, sub-Saharan Africa & Human--elephant conflict: an elephant alert e-mail opens a valve on a water supply to a shared community well \\
\bottomrule
\end{tabularx}
\end{table}

\section{Discussion}
\label{sec:discussion}

\subsection{Sovereignty Is Determined by Data Paths, Not by Where Inference Runs}

The central observation of this study is that relocating inference to a field-owned computer does not by itself keep data local. What determines whether an image leaves the site is the route by which it arrives and the route by which alerts depart. A self-hosted system that receives images through a commercial mailbox still passes every image through the mail provider, while a system that receives images by FTP/SFTP on the site network and sends alerts through a local relay need transmit nothing beyond the site. Each configuration therefore represents a different balance between convenience and control (Table~\ref{tab:options}). This framing is consistent with reviews of data sovereignty that locate its meaning in meaningful control over data, claimed by agents ranging from individuals to states~\parencite{hummel2021}: an organisation may own the computer on which inference runs without controlling the networks through which its data travel.

\begin{table}[htbp]
\centering
\caption{Deployment configurations and the parties through which images and alerts pass.}
\label{tab:options}
\footnotesize
\begin{tabularx}{\textwidth}{@{}P{2.8cm}LLLL@{}}
\toprule
Configuration & Images pass through & Alerts pass through & Set-up effort & Suited to \\
\midrule
Hosted service (e.g.\ Conservation AI, now retired) & Mail provider, then a foreign server & Foreign server, mail provider & Lowest & Sites with no local computer and no sensitive data \\
Trap Tracker RT with a cloud mailbox & Mail provider only & Mail provider & Low: a mail account & Sites where people rarely appear and species are not targeted \\
Trap Tracker RT, fully local & Nothing outside the site & Local relay; its onward route is the site's choice & Higher: FTP server, relay, network route for cameras & Protected species, people in frame, laws requiring local processing \\
\bottomrule
\end{tabularx}
\end{table}

Even the fully local configuration must eventually deliver an alert to a ranger's phone, and that final step may cross a mobile network or a mail provider. The difference is that the organisation decides what crosses the boundary, including the mail provider used for that final step; alerts that carry only a short text, leaving the images on site, are planned (Section~\ref{sec:planned}). In this respect the design gives practical effect to principles that call for data about places and people to remain under the control of those with a stake in them~\parencite{carroll2020,kukutai2016,sandbrook2021}, while leaving the organisation free to share data openly where it judges this appropriate~\parencite{wilkinson2016,tulloch2018}.

\subsection{Out-of-Distribution Accuracy and the Case for Regional Two-Stage Pipelines}

The comparison shows that the choice of model matters as much as the choice of architecture. DeepFaune-SSA maintained an accuracy of about 0.90 at sites in three countries and habitats it had never seen, whereas the three alternatives reached between 0.53 and 0.56. The way each alternative failed is instructive. The single-stage detectors, trained on the same regional data as DeepFaune-SSA, missed about a quarter of animals altogether, and replacing YOLOv10x with the newer YOLO26x made no difference. This is consistent with the literature on domain shift in camera-trap imagery~\parencite{beery2018,schneider2020,norman2023} and with the expectation that a detector trained on millions of images from many projects~\parencite{beery2019,hernandez2024} transfers to new locations more reliably than one trained on a single regional collection. Because the training data were held constant, the gap between the single-stage and two-stage models can be attributed to the architecture: decoupling detection from classification means that the regional component only has to name an animal that has already been found, and it can be trained on cropped images, which improve classification~\parencite{gadot2024}.

SpeciesNet, which shares the two-stage design, found animals almost as reliably as DeepFaune-SSA but named them far less often. When it committed to a species it was usually right, and a third of its remaining answers were correct at a coarser rank. This behaviour suits its purpose as a global model used for survey triage, where a cautious ``bovidae'' is preferable to a confident error. For alerting, however, a watch-list is defined by species, and a rolled-up label produces no alert. The comparison therefore supports regional classifiers, fine-tuned from open weights on modest regional data~\parencite{fergus2026shoulders}, as the component that a field-owned alerting system most needs, while the detector that precedes them can be shared across regions.

These results should not be over-generalised. Each model was evaluated with its released weights and default settings, and SpeciesNet's thresholds for rolling up predictions could be relaxed at the cost of more errors. Nevertheless, for alerting the asymmetry of errors is important: a missed or unnamed animal produces no alert at all, whereas a misidentified animal can still prompt a human to look. The two-stage pipeline is three to four times more expensive than the single-stage models on the CPU, but even on eight threads its throughput of 19 images per minute exceeds the rate at which a typical camera network transmits images, so in this setting the additional computation is a modest price for substantially higher recall.

\subsection{The Costs of Ownership}

Self-hosting transfers responsibilities from a central provider to the field organisation: reliable power, a computer that remains switched on, software updates, backups and security. A cellular camera on the public internet must be able to reach the site's FTP server, which in practice requires a private mobile network, a virtual private network, or a single forwarded port protected by strong credentials. The web interface lacks authentication and must remain on the local machine. These requirements are not trivial. A global survey of conservation technologists identified competition for limited funding and inadequate capacity building among the most pressing constraints on the field, with some constraints falling disproportionately on practitioners in developing economies~\parencite{speaker2022}; cost remains the most frequently cited barrier to camera trapping itself~\parencite{glover2019}; and comparative evaluations of AI platforms for camera-trap data recommend weighing usability and workflow alongside accuracy~\parencite{velez2023}. A self-hosted system is accordingly most appropriate where an organisation already maintains some local computing capability, or where a partner can provide it. Against these costs must be set the benefits: the organisation retains its data, is not exposed to the discontinuation of a central service, and can continue to operate when connectivity to the wider internet is lost. Because both the application and the regional models are openly available~\parencite{traptracker2026,fergus2026ssa}, the barrier to adoption is operational rather than financial or contractual, consistent with arguments that open release distributes control over how models are adapted and deployed~\parencite{kapoor2024}.

\subsection{Real-Time Alerting as Surveillance}

Moving inference on site addresses where data go, but not what they are used for. Real-time camera alerts are, among other things, a surveillance technology, and the social science of conservation has cautioned that such technologies can extend into the monitoring of people who live and work around protected areas and can reinforce militarised approaches to enforcement~\parencite{adams2019,duffy2019,arts2015}. People detected by the system are detected in real time, and the same capability that alerts rangers to a poacher may alert them to a resident collecting firewood. Local control does not resolve these questions, but it does place decisions about who receives alerts, which classes trigger them and how long images are retained with the organisation that is accountable to the local context. Codes of conduct for camera-trap research~\parencite{sharma2020} and principles for the responsible use of monitoring technology~\parencite{sandbrook2021} provide a starting point for the governance that should accompany any deployment.

\subsection{Limitations}

Several limitations qualify these findings. The operational results derive from bench tests on a single desktop workstation; they establish the behaviour of the software under controlled conditions but not its performance on low-power field hardware or over real cellular networks. The earlier deployments described in Section~\ref{sec:deployments} used the hosted Conservation AI platform, so they support the case for real-time alerting and for local hosting, but they do not evaluate Trap Tracker RT itself; Trap Tracker RT is now in use by five organisations, but no measurements from these deployments are reported here. The out-of-distribution evaluation covers 26 of the 33 classes and caps the number of images per species, so overall accuracy depends on the composition of the sample, and per-species recall (Figure~\ref{fig:recall}) is the more transferable measure. SpeciesNet's training data were not checked against the three evaluation datasets, and any overlap would favour SpeciesNet; the single-stage models were trained on the same data as DeepFaune-SSA, which was checked. All comparison models were evaluated with their released weights and were not retrained or recalibrated for this study.

\subsection{Future Work}
\label{sec:planned}

Four measurements are needed to move from bench evaluation to field evidence: throughput on representative low-power field laptops; end-to-end latency from shutter to alert with real cellular cameras, partitioned into camera upload, settling and scanning, inference and mail delivery; alert precision and recall for each watch-listed species across confidence thresholds, obtained by replaying the out-of-distribution datasets as image streams; and stability over multi-week operation, including restarts and loss of connectivity. Planned developments include a CPU build optimised for low-power computers, an option to send alerts as short text without images, so that the images remain on site, and delivery by SMS through a gateway for areas where e-mail is unreliable, integration with protected-area management platforms from the local computer, and a systematic evaluation of the fully local configuration at the sites where Trap Tracker RT is now running.

\section{Conclusion}
\label{sec:conclusion}

Real-time camera-trap alerting has demonstrated its value to field conservation, but hosting the inference centrally has meant transmitting sensitive images across national borders and making field operations dependent on a single provider. This study has shown that the same capability can be provided on a computer owned by the field organisation, with an ingestion and alerting path on which no image need leave the site. With DeepFaune-SSA v1.2, a regional two-stage pipeline identified sub-Saharan species with an accuracy of 0.901 on out-of-distribution images from three countries, compared with 0.530 for the global SpeciesNet classifier and 0.556 and 0.547 for single-stage YOLOv10x and YOLO26x models trained on the same regional data, and on the bench it issued alerts approximately 6~s after an upload completed, including over slow links, without requiring a graphics processor to keep pace. The system is already used by five organisations in the United Kingdom and Kenya, where alerts support grey squirrel monitoring, nest and farm-gate monitoring, farm biodiversity surveys and human--elephant conflict mitigation. The principal open questions are now operational and social rather than technical: how such systems perform on modest field hardware over real networks, how organisations sustain the responsibilities of running them, and how the alerts they produce are governed.

\section*{Author Contributions}

Conceptualization, methodology, software, validation, formal analysis, investigation, data curation, writing --- original draft preparation, and writing --- review and editing: P.F. The author has read and agreed to the submitted version of the manuscript.

\section*{Funding}

This research received no external funding.

\section*{Institutional Review Board Statement}

Not applicable. This study involved no direct handling, capture, or manipulation of live animals: all imagery was collected non-invasively by static, motion-triggered camera traps, either by the projects that contributed to the training data or by the LILA BC dataset providers; no additional data collection was undertaken for this study.

\section*{Informed Consent Statement}

Not applicable, as this study did not involve human subjects. Images selected for publication were checked to contain no people.

\section*{Data Availability Statement}

The out-of-distribution evaluation images are openly available from LILA BC under CDLA-Permissive-1.0~\parencite{lilamara,liladesertlion,lilankhotakota}. DeepFaune-SSA v1.2 and the YOLOv10x and YOLO26x models are openly available on Zenodo as ONNX models~\parencite{fergus2026ssa,fergus2026yolossa}, and the sampling, overlap-check, inference and scoring code for the out-of-distribution evaluation is in the DeepFaune-SSA repository~\parencite{deepfaunessa2026repo}. Trap Tracker RT is available on GitHub~\parencite{traptracker2026}. Apart from the Snapshot Safari images, which are openly available from LILA BC~\parencite{pardo2021,swanson2015}, the training data are not publicly available: they were shared by contributing projects under data-sharing agreements made through Conservation AI before Trap Tracker was launched, and they cannot be released because of the sensitivities and terms covered by those agreements.

\section*{Acknowledgments}

The author thanks the DeepFaune initiative, the MegaDetector and PyTorch-Wildlife team, and Meta for openly releasing the models on which this work depends; the Snapshot Safari projects, WCS Nigeria and the African partners who contributed images to the training data over the last seven years; the Biome Health Project, Desert Lion Conservation and the Nkhotakota camera-trap team, and LILA BC, for the out-of-distribution evaluation images; Chester Zoo, the Uganda Wildlife Authority, the Kenya Wildlife Service, Mount Kenya Wildlife Conservancy, Welgevonden Game Reserve and GWCT Wales for the deployments this work builds on; the Animal and Plant Health Agency, GWCT Wales, Fieldmark AI and Maes Truan Farm, the Friends of Karura Community Forest Association and Project PARC for running Trap Tracker RT; the People's Trust for Endangered Species, the British Hedgehog Preservation Society, MammalWeb, the Zoological Society of London, Natural England and Hogwatch for their use of Trap Tracker Offline; and the Conservation AI and Trap Tracker development team for the underlying infrastructure.

\section*{Conflicts of Interest}

The author declares no financial conflict of interest. Trap Tracker and the Conservation AI platform referenced in this paper are research infrastructure developed by the author's research group; Trap Tracker RT is released for research and non-commercial use only, and no commercial licence is offered. The single-stage YOLOv10x and YOLO26x baselines are models from the same group.

\printbibliography

\clearpage
\appendix
\section{Example Detections}
\label{app:examples}
\setcounter{figure}{0}
\renewcommand{\thefigure}{A\arabic{figure}}
\renewcommand{\theHfigure}{A\arabic{figure}}

\begin{figure}[p]
\centering
\includegraphics[width=\textwidth,height=0.86\textheight,keepaspectratio]{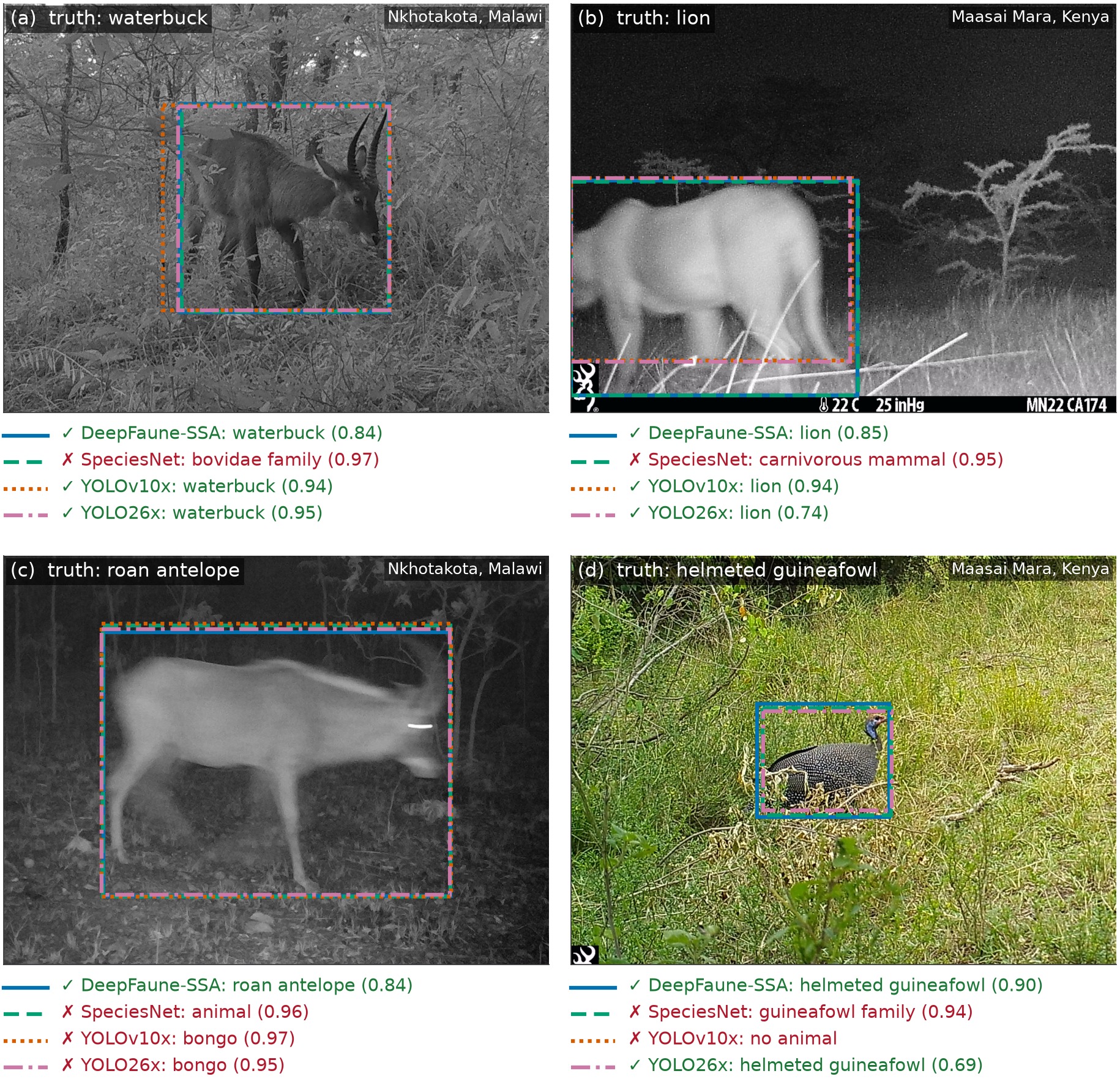}
\caption{Species-level resolution. DeepFaune-SSA v1.2 names the species, while SpeciesNet returns a coarser label: (a) waterbuck, (b) lion, (c) a motion-blurred roan antelope, also called bongo by both YOLO models, and (d) helmeted guineafowl, missed by YOLOv10x. Boxes: DeepFaune-SSA solid blue, SpeciesNet dashed green, YOLOv10x dotted orange, YOLO26x dash-dotted pink. $\checkmark$ exact species; $\approx$ correct genus only; \ding{55} neither (wrong, too coarse or no animal). Tiles are cropped for display. Images: LILA BC, CDLA-Permissive-1.0.}
\label{fig:A1}
\end{figure}

\begin{figure}[p]
\centering
\includegraphics[width=\textwidth,height=0.86\textheight,keepaspectratio]{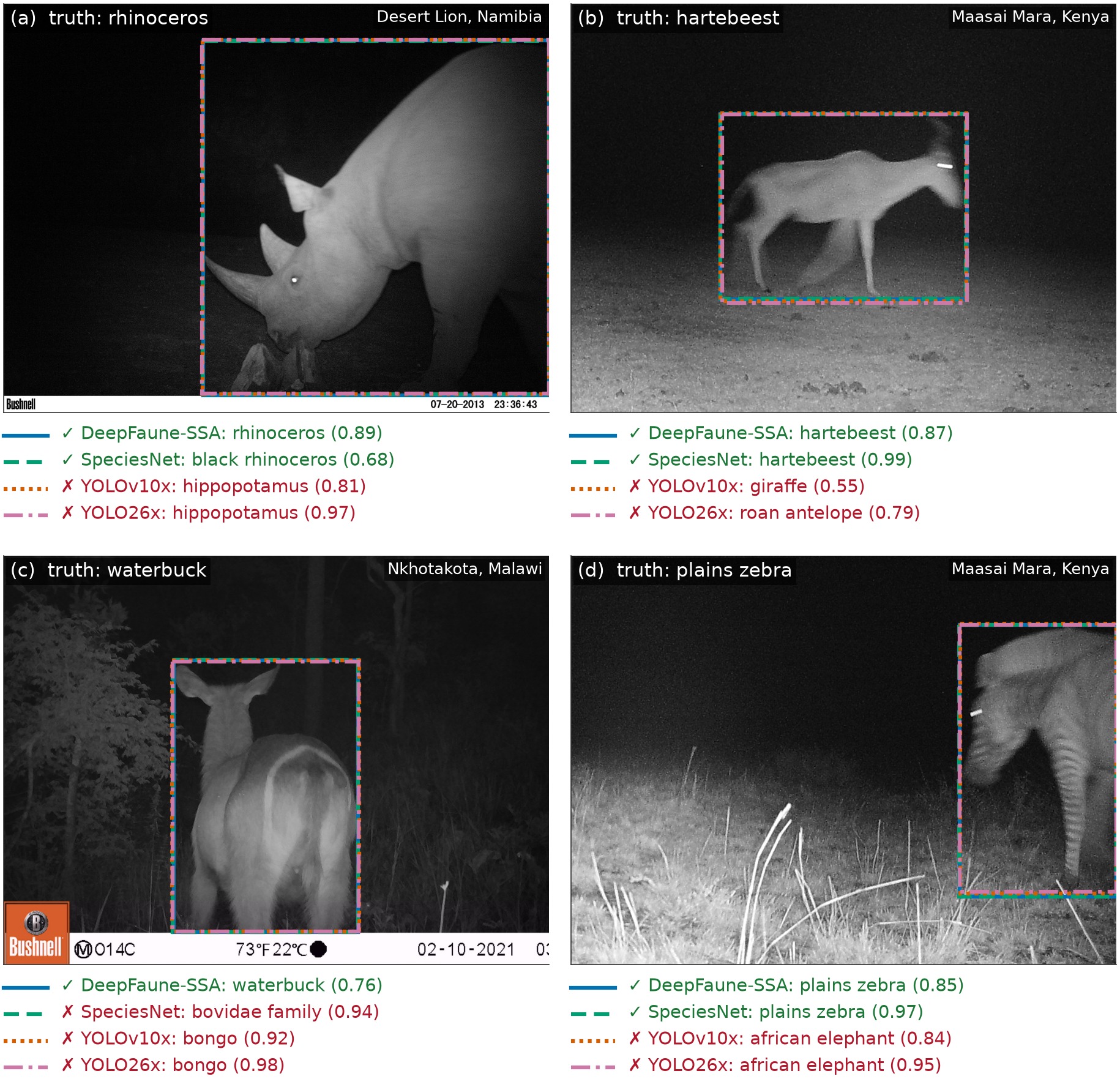}
\caption{Images on which both single-stage YOLO models find the animal but name it wrongly, while DeepFaune-SSA v1.2 names the species: (a) rhinoceros called hippopotamus, (b) hartebeest called giraffe and roan antelope, (c) waterbuck called bongo, and (d) plains zebra called African elephant. SpeciesNet is correct in (a), (b) and (d) and too coarse in (c). Boxes and symbols as in Figure~\ref{fig:A1}.}
\label{fig:A2}
\end{figure}

\begin{figure}[p]
\centering
\includegraphics[width=\textwidth,height=0.86\textheight,keepaspectratio]{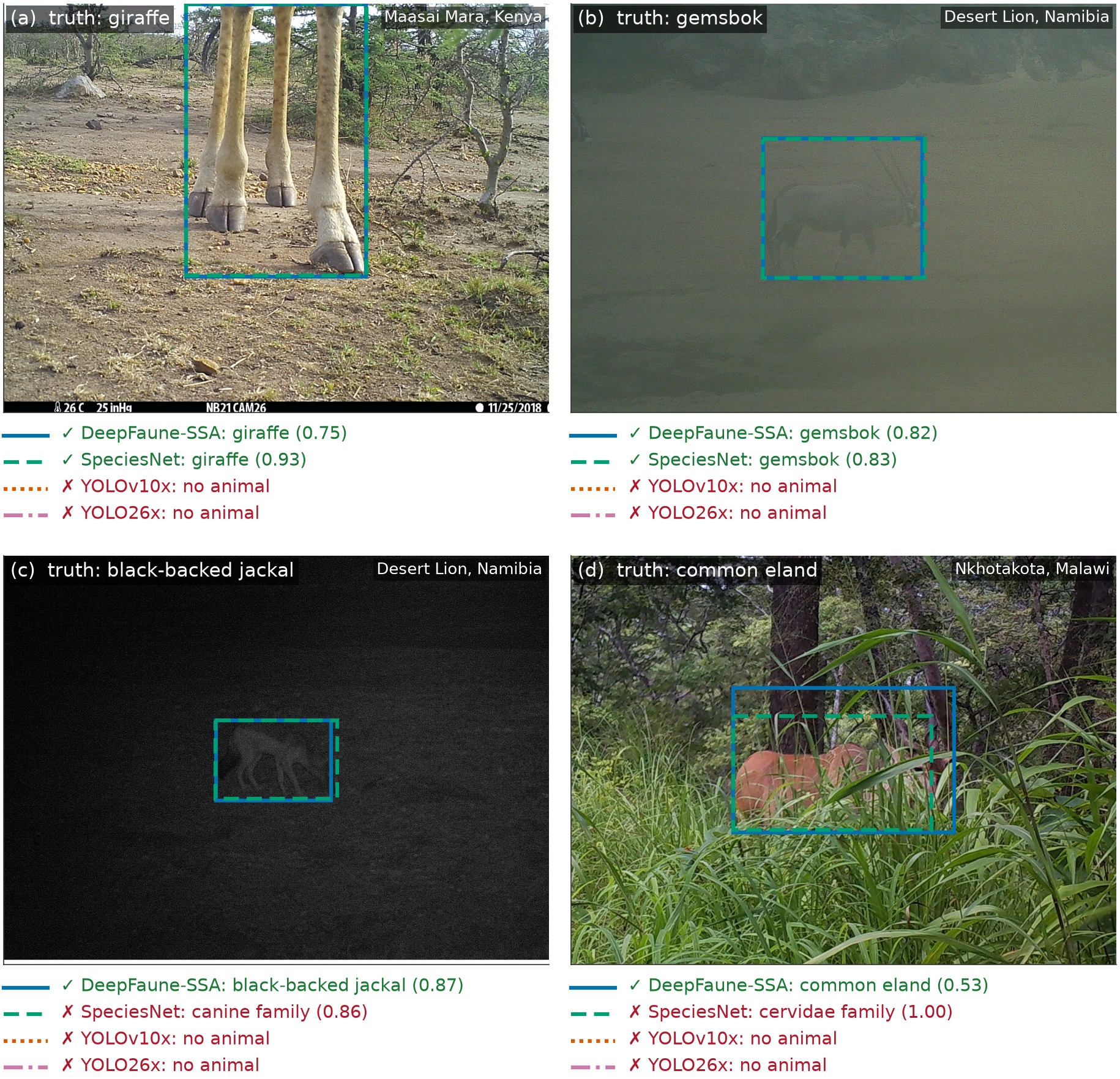}
\caption{Images on which neither single-stage YOLO model reports an animal at detector confidence 0.25, while the MegaDetector stages of DeepFaune-SSA v1.2 and SpeciesNet find it: (a) the legs of a giraffe, (b) a gemsbok in haze, (c) a black-backed jackal at night and (d) a common eland partly hidden by grass. DeepFaune-SSA names all four; SpeciesNet names (a) and (b) and returns coarser labels for (c) and (d). Boxes and symbols as in Figure~\ref{fig:A1}.}
\label{fig:A3}
\end{figure}

\begin{figure}[p]
\centering
\includegraphics[width=\textwidth,height=0.86\textheight,keepaspectratio]{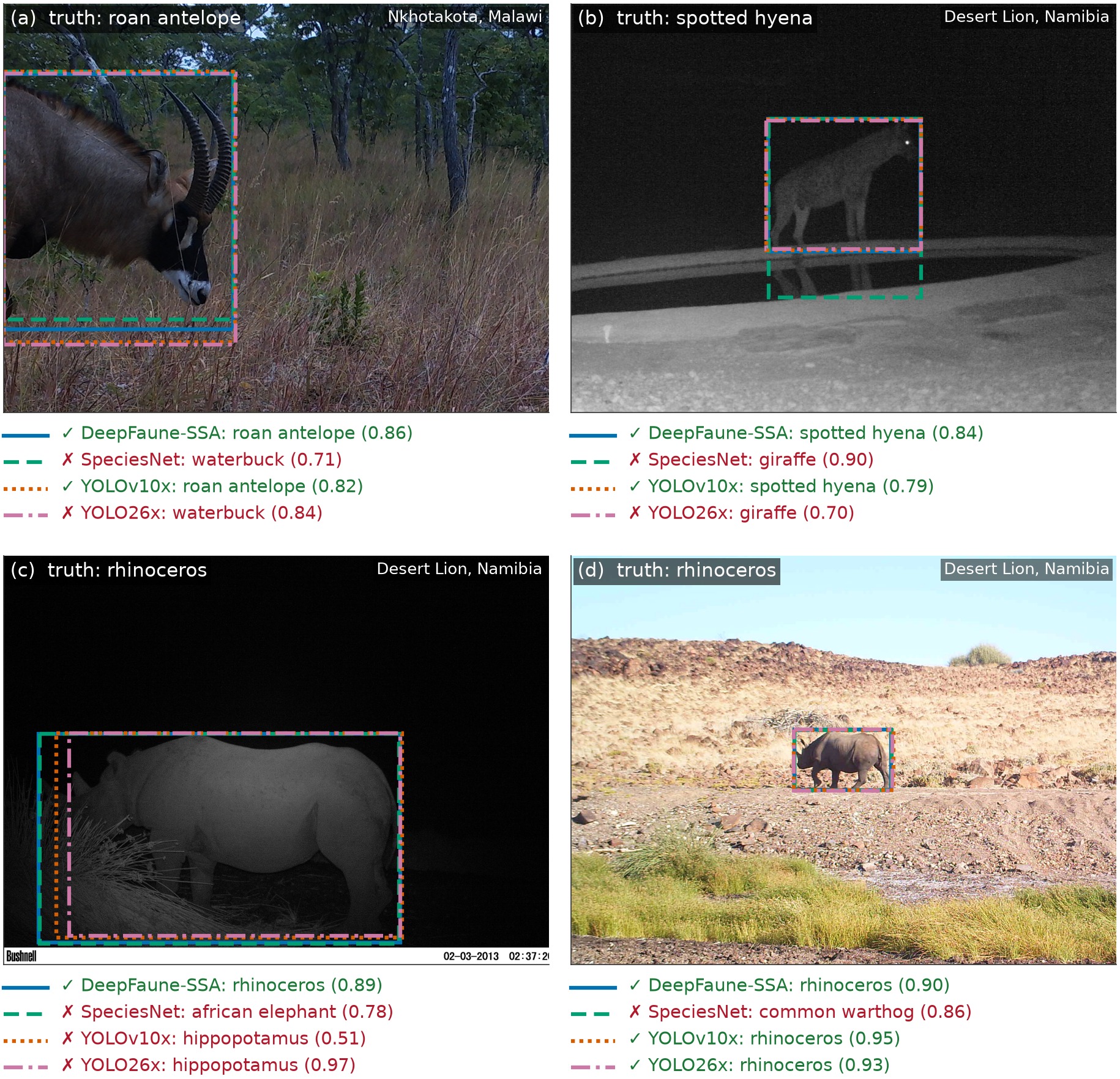}
\caption{Images on which SpeciesNet names a different species, while DeepFaune-SSA v1.2 names the correct one: (a) roan antelope called waterbuck, (b) spotted hyena called giraffe, (c) rhinoceros called African elephant and (d) rhinoceros called common warthog. YOLO26x makes the same error as SpeciesNet in (a) and (b). Boxes and symbols as in Figure~\ref{fig:A1}.}
\label{fig:A4}
\end{figure}

\begin{figure}[p]
\centering
\includegraphics[width=\textwidth,height=0.86\textheight,keepaspectratio]{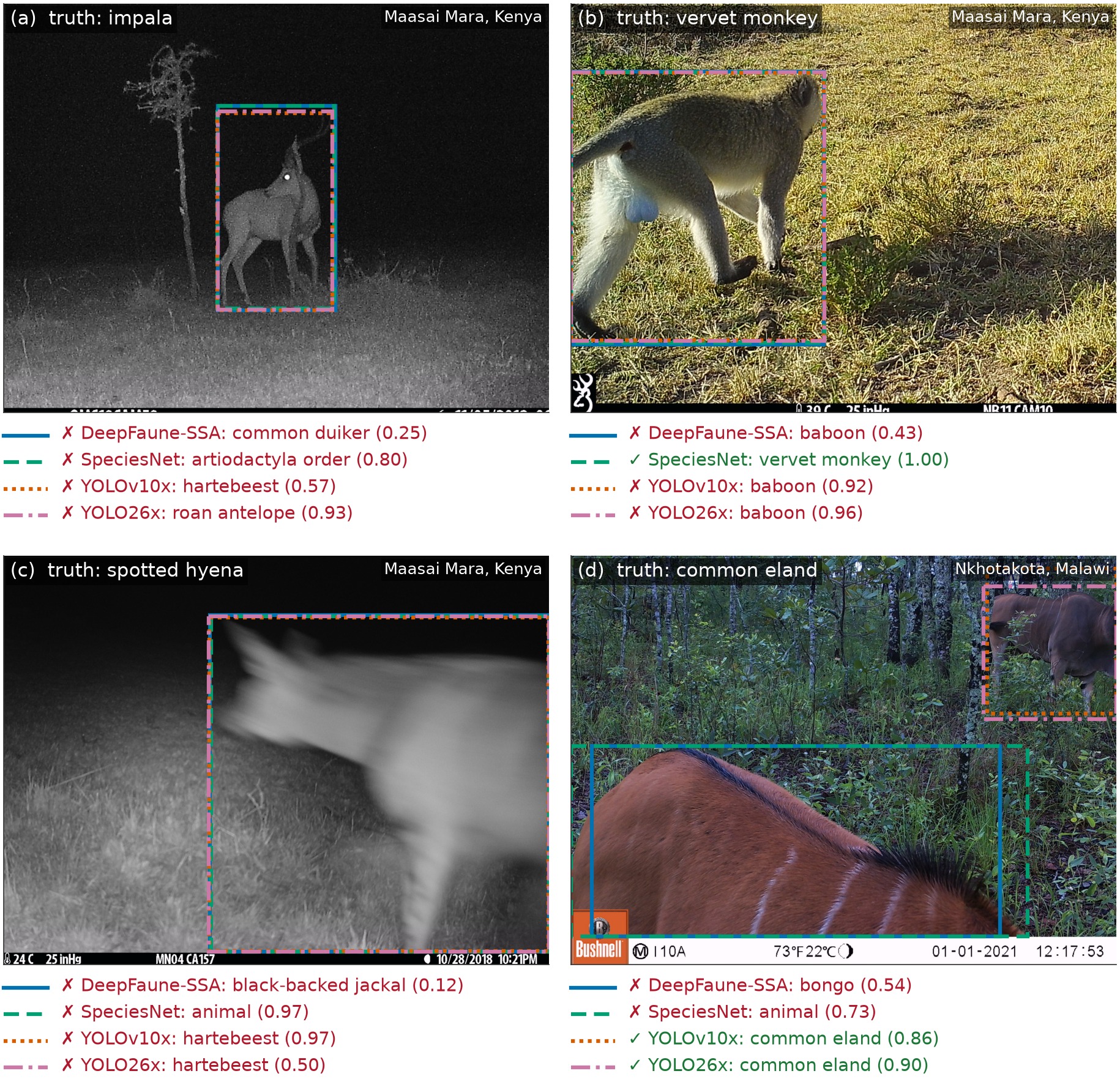}
\caption{Characteristic errors of DeepFaune-SSA v1.2, matching the commonest confusions in Section~\ref{sec:ood}: (a) an impala under infrared called common duiker, with low confidence; (b) a vervet monkey called baboon, which SpeciesNet names correctly; (c) a motion-blurred spotted hyena called black-backed jackal, with low confidence; and (d) a common eland close to the camera called bongo, while both YOLO models find and name a second eland further away. Boxes and symbols as in Figure~\ref{fig:A1}.}
\label{fig:A5}
\end{figure}

Figures~\ref{fig:A1}--\ref{fig:A5} compare the four models on images from the out-of-distribution datasets. Each tile shows the box behind each model's image-level label, that is, its most confident animal box with detector confidence of at least 0.25 (Section~\ref{sec:ood-methods}), drawn in the model's colour and line style: DeepFaune-SSA v1.2 solid blue, SpeciesNet dashed green, YOLOv10x dotted orange and YOLO26x dash-dotted pink. Where boxes coincide they are drawn over one another. The box shown for SpeciesNet is its MegaDetector v5a detection, but its label is the ensemble prediction for the whole image. Below each image are the four labels with their confidences (for DeepFaune-SSA the classifier's, for the YOLO models the detector's and for SpeciesNet the ensemble score). DeepFaune-SSA confidences rarely exceed about 0.9 because its classifier was trained with label smoothing of 0.1. Labels are marked $\checkmark$ when they name the exact species, $\approx$ when they are correct only at genus level, and \ding{55} when they match neither, which includes coarser labels such as ``bovidae family'' or ``animal'' and images in which no animal was found. These are the strict and genus-level criteria of Section~\ref{sec:ood-methods}; images labelled only to genus in the source dataset, such as vervet monkey, are marked at genus level.

The images were chosen by hand for legibility from pools selected automatically from the per-image results (for example, images that DeepFaune-SSA labelled correctly and both YOLO models did not), and were checked to contain no people. They illustrate the patterns quantified in Sections~\ref{sec:ood} and~\ref{sec:comparison} rather than estimate their frequency. Tiles are cropped around the boxes for display; all models processed the full frame. Images: LILA BC, CDLA-Permissive-1.0~\parencite{lilamara,liladesertlion,lilankhotakota}.

Figure~\ref{fig:A1} shows the commonest SpeciesNet outcome after a correct species: a label that is correct but too coarse to trigger a species alert, such as ``bovidae family'', ``carnivorous mammal'' or ``animal''. Figure~\ref{fig:A2} shows the single-stage models' naming errors: both YOLO models find the animal, with boxes almost identical to those of the other models, but name a species of similar build or colour, such as hippopotamus for rhinoceros or bongo for waterbuck. Figure~\ref{fig:A3} shows their detection failures, in which neither YOLO model reports an animal at the 0.25 threshold although the MegaDetector stages of DeepFaune-SSA and SpeciesNet find it: partial views, haze, darkness and vegetation. Figure~\ref{fig:A4} shows the less common case in which SpeciesNet names a wrong species, as it did for rhinoceros in the Desert Lion images, which it called African elephant, buffalo or warthog. Figure~\ref{fig:A5} shows characteristic errors of DeepFaune-SSA, matching the commonest confusions in Section~\ref{sec:ood}. No model names the impala in (a) or the hyena in (c); SpeciesNet names the vervet monkey in (b), and in (d) both YOLO models name the more distant eland correctly.

\end{document}